# Clustering with Temporal Constraints on Spatio-Temporal Data of Human Mobility

Yunlong Wang[1], Bjoern Sommer[2], Falk Schreiber[2], Harald Reiterer[1]

[1]HCI Group, University of Konstanz, Konstanz, Germany
[2]Computational Life Sciences Group, University of Konstanz, Konstanz, German
`{firstname.lastname}@uni-konstanz.de`

**Abstract.** Extracting significant places or places of interest (POIs) using individuals' spatio-temporal data is of fundamental importance for human mobility analysis. Classical clustering methods have been used in prior work for detecting POIs, but without considering temporal constraints. Usually, the involved parameters for clustering are difficult to determine, e.g., the optimal cluster number in hierarchical clustering. Currently, researchers either choose heuristic values or use spatial distance-based optimization to determine an appropriate parameter set. We argue that existing research does not optimally address temporal information and thus leaves much room for improvement. Considering temporal constraints in human mobility, we introduce an effective clustering approach – namely POI clustering with temporal constraints (PC-TC) – to extract POIs from spatio-temporal data of human mobility. Following human mobility nature in modern society, our approach aims to extract both global POIs (e.g., workplace or university) and local POIs (e.g., library, lab, and canteen). Based on two publicly available datasets including 193 individuals, our evaluation results show that PC-TC has much potential for next place prediction in terms of granularity (i.e., the number of extracted POIs) and predictability.

## 1 Introduction

Chronic conditions, such as cardiovascular diseases and diabetes, place a heavy burden on individuals and society. Unhealthy lifestyle choices like smoking, physical inactivity, and unbalanced food intake are highly related to these chronic diseases. Therefore, an increasing number of studies are focusing on personalized health behavior change to help individuals prevent the onset of chronic diseases [1, 2]. For instance, the *just-in-time adaptive interventions (JITAIs)* framework emphasizes context detection to provide personalized interventions for behavior change [3]. In this paper, we focus on extracting *places of interest* (*POIs*, see Sect. 3.2 for a formal definition) of human mobility, since they have the potential to reflect human habits and lifestyles. The identified patterns for POIs can support the better understanding of human behavior and support increasingly personalized interventions for promoting health. In recent studies, POI-based interventions have shown great potential in mobile health applications [4, 5]. For example, MyBehavior [4] generates personalized activity suggestions tailored to different places according to the activity level of users.

In many human mobility analysis systems, the extraction of POIs is an indispensable component, while spatio-temporal data clustering is the core technique for POI extraction [6–10]. Classical clustering techniques include hierarchical clustering (e.g., Linkage [11]), partitional clustering (e.g., k-means [12]), grid-based clustering [13], and density-based clustering (e.g., DBSCAN [14]). One challenge to apply these clustering techniques in the domain of human mobility analysis is to determine the most suitable parameters, which are usually chosen heuristically (e.g., 50 meters as the minimum distance between POIs) or by using optimization techniques (e.g., Silhouette Coefficient [15], Davis-Bouldin index [16], or the reachability-based method [17]). As an optimization problem, the typical objective in clustering is to obtain high intra-cluster similarity and low inter-cluster similarity [18]. For spatio-temporal data of human mobility, however, only spatial information is considered, while temporal information is often neglected [7, 10, 19]. In this paper, with consideration of two temporal constraints in human mobility data (i.e., visit frequency and duration of visit for POIs), we will provide an effective method for estimating the optimal cluster number in hierarchical clustering for extracting POIs from spatio-temporal data of human mobility.

POI clustering approaches are often discussed in the context of next place prediction [9, 19, 20]. The proposed approach in this paper is also aimed to provide high-quality input data for next place prediction. Therefore, we adopt a metric called *predictability limit* [21] to evaluate the extracted POI sequences by different approaches.

The remainder of this paper is organized as follows: The next section discusses related works. In Sect. 3, the details of the proposed approach will be presented, including the definitions of POI and our POI score, as well as our POI extraction algorithm. Sect. 4 presents the experimental results of the proposed approach compared to four other approaches from related works. Following, in Sect. 5, we discuss the contributions and limitations of this work. The final section concludes the paper and points out the potential future work.

## 2 Related Work

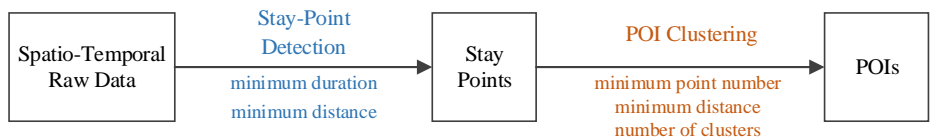

**Fig. 1.** The workflow adopted in related works [7, 9, 10] .

In the field of spatio-temporal data clustering of human mobility, the workflow shown in Fig. 1 is typically used in relevant approaches (e.g., [7, 9, 10]). It contains three states (i.e., raw data, stay-points, and POIs) and two steps (i.e., stay-point detection and POI clustering). In the first step of the workflow, a stay-point detection algorithm is used based on the approach proposed by Ye et al. [7]. A stay-point is defined as a region where an individual spends at least a predefined period of time within a certain distance. Because all data points are sequentially processed, multiple stay-points generated in the first step may be closely located and belong to one cluster. Therefore, in the second step, a clustering algorithm is used to obtain POIs.

Different parameters are required depending on the adopted clustering algorithm. For example, in the case of density-based clustering [7–9], two parameters should be set as the constraints: the minimum data points in a cluster and the reachability distance threshold [14]. These parameters determine how a POI is defined. For example, in [7] the authors roughly regard a POI as a geographic area in which a) the user stayed more than 30 minutes per visit, b) the user visited more than four times in the user's overall mobility history, and c) any distance between two data points is less than 200 meters within the geographic area. However, the reason why these values were chosen was not explained. Researchers typically set the parameter based on their own experience (e.g., [10]). In other cases, they did not report the used parameters at all [8].

**Table 1.** Summary of parameters influencing POI definitions in the related approaches. The dashes depict unused parameters in the corresponding approaches. Categories: Blue: Stay-point detection, Orange: POI Clustering.

| | Minimum duration $\Delta$ | Maximum distance $\theta$ | Reachability distance $\varepsilon$ (DBSCAN) | Minimum points *MinPts* (DBSCAN) | Visit frequency | Duration of visit | Cluster number optimization method |
|---|---|---|---|---|---|---|---|
| [7] | 30 minutes | 200 meters | 150 meters | 4 | - | - | - |
| [10] | 30 minutes | 250 meters | 250 meters | 2 | - | - | - |
| [19] | 20 minutes | 50 meters | - | - | - | - | SC, DB, semantic similarity |
| [8] | - | - | Not reported | Not reported | - | - | - |
| Our Method | - | - | - | - | $F_{vd}$ | $D_{vd}$ | *POI score* |

Different parameters used among the mentioned works are listed in Table 1. In terms of the temporal information used in the listed approaches, besides the minimum duration for a cluster, the minimum data points in a cluster can be understood as the overall visit times. The used thresholds of these temporal constraints were not clearly related to human routines (e.g., daily or weekly visits).

In [19] the authors also pointed out the challenge of choosing the appropriate parameters. Hierarchical clustering was used in the aforementioned work. To get the optimal cluster number, four metrics were compared (Silhouette coefficient [15], Davies-Bouldin index [16], and their own combinations with a semantic score, defined as the semantic similarity of the data points in a cluster). However, temporal information was ignored for clustering. Additionally, the hierarchical structure was not taken into account throughout the process of POI extraction. For example, a relatively large POI (e.g., a university) contains several smaller POIs (e.g., the library and the canteen of the university).

Therefore, we designed our approach based on three considerations. The approach should:

(1)    involve both spatial and temporal information for clustering,

(2)    gain the best granularity based on sophisticated metrics while considering the hierarchical structure among POIs, and

(3)    allow for an easy adjustment of the predefined parameters, which should be comprehensible and flexible.

## 3    Approach to Extracting POIs

In this section, we will discuss the workflow in our POI extraction approach, namely *POI clustering with temporal constraints (PC-TC)*, as shown in Fig. 2. Our approach contains four states and three steps. The first step, data pre-processing, will be discussed in Sect. 4.2, since this step depends on the properties of the dataset. This section presents the core of our approach, including the temporal information on human mobility, the definitions of POI and our POI score, and our POI extraction algorithm.

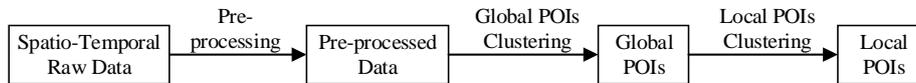

**Fig. 2.** The workflow of our approach. It contains three steps: data pre-processing, global POIs clustering, and local POIs clustering.

### 3.1    Temporal Information in Human Mobility

Human mobility, especially for office workers or students, follows schedules that are normally stable and periodic. Indeed, most humans are creatures of habit and follow well-established daily routines. The 2D map in Fig. 3 shows the locations visited and transitions of two students in Dartmouth College [22]. We can easily recognize some clusters (e.g., the ones represented as black circles). Meanwhile, we may also find it difficult to cluster some places with sparse location data exclusively based on distance (e.g., the one in the black rectangle). However, in the 3D map in Fig. 3, the identification of the corresponding cluster was improved by showing the transitions over time. The places in the black rectangle can be separated by temporal patterns (i.e., visiting frequency and duration). This is the case because the cluster marked by the rectangle was clearly more frequently visited than other clusters in the surrounding region. This example shows that additionally exploring the temporal information is crucial to understanding and analyzing human mobility patterns, especially for detecting POIs.

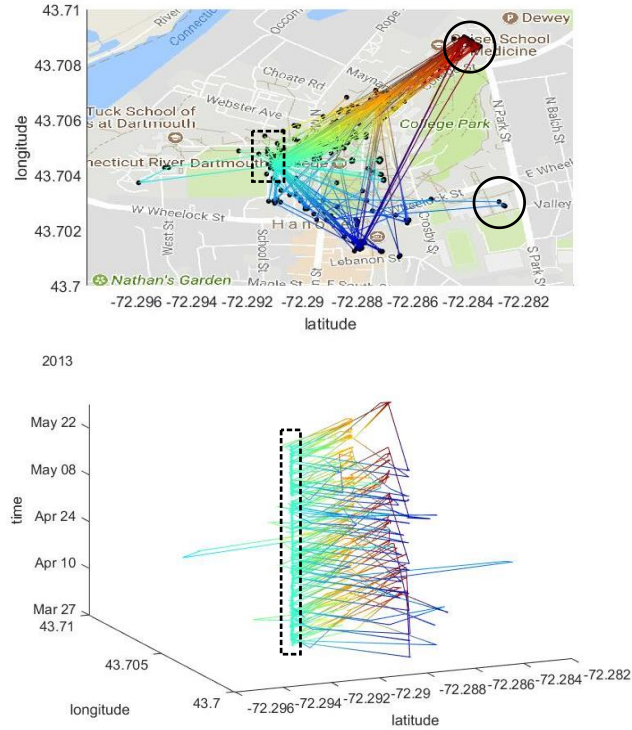

**Fig. 3.** Location data points and transitions of two college students. In the 2D map (the upper one), the black dots represent data points of location and the lines show the transitions. The black circles demonstrate two spatially distinct clusters while the black rectangle highlights a potential cluster that cannot be clearly recognized using only spatial information. By adding the time dimension, the black rectangle in the 3D figure (the lower one) shows the significant visit frequency of the location in the black rectangle in the 2D map.

### 3.2 Definitions of POIs

Based on the definitions in [10], we define POIs by taking into consideration duration of visit, visit frequency, and spatial distance. In our approach, a *POI (place of interest)* refers to a geographic area where a person (1) stays for a period of time longer than a threshold per visit day on average (*THOLD_D*, e.g., 30 minutes per visit day) and (2) visits in amount of days above a threshold (*THOLD_F*, e.g., 3 days per week). That is, the definition of the POIs relies significantly on the duration, visit frequency, and space, rather than exclusively focusing on spatial distance. We emphasize the visit days and durations per visit day instead of visit times and durations per times in order to filter out noise. For example, a person may visit some places a number of times but only on a single day in case of accidents or emergencies. It should be noted that we do not set a spatial limitation as to achieve the best spatial granularity.

As mentioned in the introduction, the number of a person's POIs in real life is small (i.e., several to dozens), normally including home, workplace or school, shopping locations (e.g., malls or supermarkets), and several other places (e.g., cinema and hospital). Some POIs, especially workplace or school, where people spend most of their waking time, may contain multiple smaller significant places such as lab, office, canteen, library, or even a rest place in the garden the user likes to visit. In order to meet our requirement to keep the hierarchical structure among POIs, we define two types of POIs including *global POIs* and *local POIs* in two spatio-temporal tiers as shown in the POI tree in Fig. 4. One global POI normally contains several local POIs (e.g., global POI 3 in Fig. 4). For example, Max goes to the university (global POI) to work, where he visits his office (local POI), the laboratory (local POI), and the campus coffee shop (local POI) on weekdays. There could also be global POIs containing no local POI (e.g., global POI 2 in Fig. 4). Other clusters in the layer of global POIs refer to the clusters not meeting the requirement of global POIs. However, these clusters may contain special local POIs without belonging to any global POI.

We argue that detecting global POIs is important, because: (1) visits to a global POIs should be more frequent than the subordinate local POIs, which may provide more predictability, and (2) detecting global POIs can enable more in-situ interventions (e.g., providing route suggestions among local POIs to coverage more walking if a user is approaching the corresponding global POI).

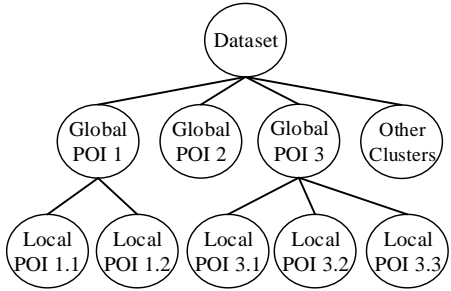

**Fig. 4.** The workflow of our approach. It contains three steps: data pre-processing, global POIs clustering, and local POIs clustering.

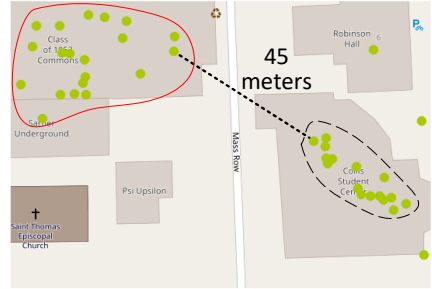

**Fig. 5.** Assume the minimum distance between the two potential POIs in two buildings is 45 meters, then a stay-point detection or DBSCAN with a distance threshold of 50 meters cannot divide the two clusters. The green color indicates locations belong to one POI resulted from DBSCAN.

To sum up, the temporal constrains for POIs include the frequency of visit days ($F_{vd}$) and the average duration per visit day ($D_{vd}$). Now we show the formal definitions of global POIs and local POIs with the corresponding thresholds for every cluster $c$:

$$\forall c \in \{global\ POIs\}, F_{vd}(c) \geq THOLD\_F_g \ and \ D_{vd}(c) \geq THOLD\_D_g \quad (1)$$

$$\forall c \in \{local\ POIs\}, F_{vd}(c) \geq THOLD\_F_l \ and \ D_{vd}(c) \geq THOLD\_D_l \quad (2)$$

### 3.3 POI Clustering Algorithm

Given the POI definition, we now show the details of our POI clustering algorithm. To obtain the optimal spatial granularity, we choose hierarchical clustering as the foundation of our approach. Another reason for choosing hierarchical clustering is that it allows the maintenance of the hierarchical structure of clusters. Thus, we can detect global POIs and local POIs in hierarchical order. The key is to find the optimal scale, that is, the optimal number of clusters. Instead of using spatial distance-based metrics [19], we use a new metric called *POI score*, which is the number of POIs among the given clusters, to evaluate the clustering result. A greater POI score means more extracted POIs. To find the optimal cluster number, we only need to try all possible cluster numbers based on the hierarchical clustering to find the number that generates the most POIs.

To illustrate how the POI score can help optimize hierarchical clustering, we use an example with real-world locations as shown in Fig. 5. Assume that the minimum distance between two potential POIs in two buildings is 45 meters. If the stay-point detection algorithm or DBSCAN is used to cluster the shown data with a distance threshold of 50 meters, the locations will be in one cluster. Using our approach, we check if more POIs can be generated when we divide them into two clusters or more according to their hierarchical order. In this example, we finally get two POIs (marked with red or black area).

The process of using the POI score based on the temporal constraints of POIs to find the optimal cluster number of the hierarchical clustering is expressed in Formula 3. The function $POI\_SCORE(C_n, P)$ is to calculate the POI score given the POI parameters $P$ and the clusters $C_n$, where $n$ ranges from 2 to the data size $N$.

$$\underset{2 \le n \le N}{\arg\max} \, POI\_SCORE(C_n, P) \tag{3}$$

$$subject\ to: POI\ Definition \qquad \text{(See Formula 1 and 2)}$$

Algorithm 1 demonstrates the procedure of extracting global POIs. Given the prepared trajectory and the parameters for POI determination, the hierarchical clustering (C-Linkage [23]) is firstly applied (Algorithm 1, row 2). Next, the clustering results with cluster number $n$ from 2 to $N$ are evaluated using the POI score. Normally, the POI score increases at the beginning of the iterations, when the cluster number is smaller than the actual maximum of POI score, while it decreases when the cluster number becomes too large. In the rare and extreme case that every data point represents a cluster, no cluster will likely meet the temporal constraints. Therefore, we detect several conditions where the loop can be terminated to avoid unnecessary computation. These conditions include: (1) the POI score equals zero, (2) the POI score does not increase for more than 50 iterations, and (3) the POI score decreases to be smaller than the highest thus far by 10. This strategy saves computational resources when the data size is large.

After the optimal cluster number is obtained, it is used to generate the clustering result. The clusters that meet the constraints of global POI will be used for clustering local POIs, while the other clusters will be checked if they meet the constraints of local POI (Algorithm 1, row 14). The process for local POIs is the same but with the constraints for local POI (Algorithm 1, row 6).

---

**Algorithm 1: Extracting Global POIs**

INPUT: The trajectory T; Thresholds $THOLD_g$ and $THOLD_l$ for global and local POIs, respectively.

UOTPUT: The set of POIs G.

1: $G \leftarrow \{\}$;
2: $L \leftarrow$ LINKAGE(T);
3: Score $\leftarrow \{\}$;
4: FOR $n = 2 : N$
5:      $C_n \leftarrow$ CLUSTER(L, n);
6:      $Score_n \leftarrow$ POI_SCORE($C_n$, $THOLD_g$);
7:      IF MeetTerminateConditions
8:          BREAK;
9:      END
10: END
11: BestClusterNumber $\leftarrow$ ARG MAX(Score);
12: $C \leftarrow$ CLUSTER(L, BestClusterNumber);
13: $G \leftarrow$ POI_SCORE(C, $THOLD_g$);
14: $G \leftarrow G \cup$ POI_SCORE(C, $THOLD_l$);
15: RETURN G

---

The worst-case computation complexity of searching for the maximum POI score is $O(n^2)$. As the computation complexity of the adopted hierarchical clustering algorithm (Algorithm 1, row 2) is $O(n^2)$, the total complexity of Algorithm 1 is $O(n^2)$.

## 4 Evaluation

### 4.1 Datasets

We made use of two publicly available datasets for our evaluation. The first dataset stems from the StudentLife Study [22]: the dataset contains heterogeneous data of 49 students collected from a class of Dartmouth College in the U.S. for around 10 weeks in 2013. The GPS location data was collected every 10 or 20 minutes by a dedicated application on participants' own Android smartphones. The other dataset comes from the Mobile Data Challenge (MDC) [24, 25], which was collected from Oct. 2009 to Mar. 2011 and involved around 200 people. The data was collected by a dedicated application using Nokia N95 and most of the participants were employees and college students. Unlike the first dataset, the time interval of data collection is from 1 minute to 10 minutes according to the state of the phone [24]. We show more details of the two datasets in our supplementary material.

The data collection duration in the MDC dataset (18 months) is much longer than the one in the StudentLife dataset (10 weeks). To make the MDC dataset comparable to the StudentLife dataset, we exclusively selected the participants' data from the MDC dataset covering more than 60 days, which resulted in a dataset containing the data of 144 participants. For this dataset, we only keep 60 days' data for each participant.

## 4.2 Data Pre-processing

Since all human mobility data was collected during the users' real-life activities, erroneous data or missing data exists in both datasets. Any missing data must thus be detected in the pre-processing step, because it is related to the duration estimation. Since both datasets have a minimum time interval for data collection, a time interval of 30 minutes was used to identify missing data. We regard the location data with a confidence radius larger than 1,000 meters or equal to 0 as erroneous data.

After labeling missing data and removing error data, the continuously un-changed location data are deleted while the first valid location data are retained and the durations are calculated.

## 4.3 Compared Approaches and Parameters

Resulting from our review of related works, we selected four other approaches to compare with our approach (see Table 2). All compared approaches follow the workflow shown in Fig. 1. Approach 1 and 2 use density-based clustering (OPTICS [26] and DBSCAN [14]). Compared to DBSCAN, OPTICS only need one parameter (the minimum point number *MinPts*) [17]. The implementation of the OPTICS approach is adapted from [27], while the DBSCAN approach is implemented based on [28]. Approach 3 and 4 are based on complete-linkage (C-Linkage) algorithm [23], where *Davies-Bouldin (DB)* index [16] and *Silhouette coefficient (SC)* [15] are used to evaluate the generated clusters and find the optimal number of clusters, respectively. The built-in functions in Matlab related to C-Linkage, DB, and SC are used in corresponding approaches. In the stay-point detection, we adopt the algorithm described in [7], with one difference that we avoid involving missing data points into stay points. Thus, the accumulation of the duration for missing data points is avoided.

**Table 2.** The list of different approaches and parameters in the experiments. SP means stay-point detection. The $N$ in *MinPts* means the data size. The other symbols have the same meaning as in Table 1.

| No. | Approach | Abbr. | $\Delta$ (minutes) | $\theta$ (meters) | $\varepsilon$ (meters) | *MinPts* |
|---|---|---|---|---|---|---|
| 1 | SP + OPTICS | OPTICS | 30 | 50 | - | $\log_{10}(N)$ |
| 2 | SP + DBSCAN | DBSCAN | 30 | 50 | 50 | $\log_{10}(N)$ |
| 3 | SP + C-Linkage(DB) | DB | 30 | 50 | - | - |
| 4 | SP + C-Linkage(SC) | SC | 30 | 50 | - | - |
| 5 | PC-TC | PC-TC | - | - | - | - |

It should be noticed that we use a technique to simplify the distance computation using GPS data when the absolute distance is not necessary. Among the five approaches, only DBSCAN needs the absolute distances except stay-point detection. Therefore, we apply the technique to all the other approaches. Please refer to the supplementary material for the detail of our distance computing method.

We conducted two evaluations using different parameter settings. Firstly, we use all the listed methods with a set of heuristic parameter values. Secondly, we only compare our method with DBSCAN by choosing several values for the key parameters in a reasonable range.

All tested approaches were implemented in Matlab 2017a and the evaluation was run on a Lenovo ThinkPad laptop with 8 GB memory and Intel i7-5600U CPU (2.6 GHz).

**First evaluation.** We use 50 meters [7, 10] and 30 minutes [19] as the corresponding parameters in stay-point detection. The parameter *MinPts* is set to $\log_{10}(N)$ as recommended in [29] in OPTICS and DBSCAN. The distance parameter in DBSCAN is also set to 50 meters, the same as the distance parameter in stay-point detection. In our approach, we set the parameters under office working scenario as following. For a global POI (e.g., workplace), we assume that a user visits the location every day during weekdays. Thus, the threshold for $F_{vd}$ should be around 5/7. Considering an exception rate of 10% (e.g., traveling somewhere else), we suggest using 0.63 as the threshold for $F_{vd}$ of global POIs. Similarly, we assume that a user visits a local POI at least once a week and suggest using 0.13 (1/7 * 90%) for local POIs. In terms of the $D_{vd}$, we use 120 minutes as the threshold for global POIs, and 30 minutes for local POIs, the same as in stay-point detection.

**Second evaluation.** To investigate the effect of parameters in PC-TC, we need to conduct another evaluation. We only compare our approach with DBSCAN, which is the most used approach in related works. In TC-PC, we set $F_{vd}$ of local POIs to a range of values (0.1-0.9). These values cover visit patterns from weekly to daily. In DBSCAN, we use 4 values (1, 5, 25, and 50 meters) for the distance parameter. The other parameters remain the values in the first evaluation.

### 4.4 Evaluation Metrics

We evaluated the chosen approaches based on three metrics, namely, numbers of POIs, predictability limit [9, 21] of POIs, and computation time. The number of POIs roughly indicates how much information or patterns were extracted, the predictability limit shows the quality of the POIs in terms of next place prediction, while the computation time demonstrates the approaches' efficiency.

The limits of predictability in human mobility was initially investigated using cellular-scale location data by Song et al. [21]. The upper limit of human mobility predictability represents the average probability of correctly predict the next POI by a proper algorithm, given a POI sequence. We use *predictability limit (PL)* to refer to the theoretical upper bound of predictability in the rest of the paper. We show the method of calculating the PL [21] in our supplementary material. Naturally, a larger number of POIs provides more mobility information, but it also potentially increases its entropy rate. Greater entropy rate of a sequence can increase the prediction uncertainty and decrease the PL [9]. Our method is designed to explore the most POIs with temporal constraints. On the other hand, the temporal constraints can avoid places below a frequency threshold (e.g., a restaurant visited only once). Thus, it may decrease the entropy rate and increase the predictability limit. We want to explore how our method can balance this tradeoff compared to other methods. To the best of our knowledge, this is the first time PL is used to evaluate POI clustering results.

### 4.5 Results

**First evaluation.** The results of the first evaluation are shown in Fig. 6. The compared approaches define no such temporal constraints for POI-determination as we define in our approach. Thus, we regard all generated clusters in the other approaches as POIs. The numbers of POIs are shown in the two left most graphs in Fig. 6. For the StudentLife dataset, our approach (PC-TC) generates significantly more POIs than other approaches, while for the MDC dataset our approach yields less POIs than DB and SC.

The StudentLife dataset only contains data from students in Dartmouth College, which is a self-contained campus [22]. Therefore, most of the participants' location data were generated on campus. The potential POIs, such as the dormitory, the classrooms, and the library, are close to each other. Referring to the example in Fig. 5, POIs with a distance of less than 50 meters (the threshold for stay-points) will not be separated in the compared approaches. In PC-TC, there are no constraints for spatial distance, which is why PC-TC obtains more POIs. Unlike the dataset representative of campus life, the visited places of the participants in the MDC datasets are generally farther apart from each other. We observed that many participants in this dataset visited several cities with low frequency in Switzerland. In PC-TC, the places which cannot meet the temporal constraints are filtered out, which explains why the numbers of POIs in PC-TC for the MDC dataset do not exceed the ones of other approaches as for the StudentLife dataset. By contrast, DB and SC generate much more POIs because they have no constraints on visit duration or frequency of POIs.

In addition to the total number of local POIs in PC-TC, we also exclusively show the number of global POIs. The numbers of global POIs vary from one to five for all participants of both datasets, which is reasonable in the real world.

As discussed in Sect. 4.4, there is a trade-off between POI number and predictability limit (PL). The two graphs the center of Fig. 6 show that for the StudentLife dataset our approach generates more POIs but results in a lower PL (0.67 in the median). However, with much larger POI numbers, our approach obtains nearly the same PL as OPTICS (0.71 in the median) for the StudentLife dataset. Meanwhile, for the MDC dataset, with very similar POI numbers (6-10 in the median), our approach achieves a higher PL (0.92 in the median) than OPTICS (0.76 in the median) and DBSCAN (0.83 in the median). This demonstrates the potential of our approach of balancing the tradeoff between POI number and PL.

Since the numbers of global POIs are very small, the predictability limit levels are always higher than for local POIs. Therefore, global POI sequences provide an alternative in next place prediction tasks.

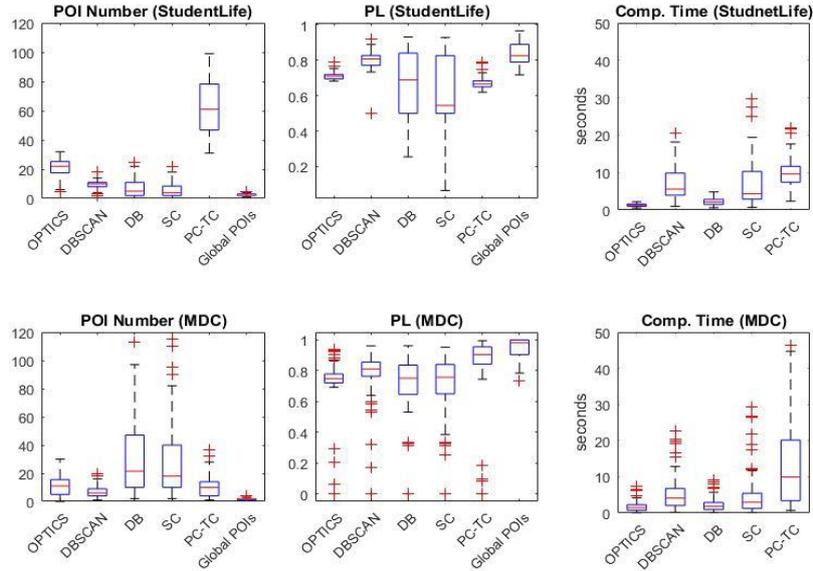

**Fig. 6.** The evaluation results of different approaches for the StudentLife dataset and the MDC dataset. The two left most graphs show the numbers of extracted POIs; the two middle graphs present the predictability limit of the POI sequences; and the two right most graphs represent the computation time.

In terms of computation time, shown in the right most graphs in Fig. 6, our approach requires more time than the other approaches. This result can be explained by two reasons. First, the computation complexity of PC-TC ($O(n^2)$) is higher than DBSCAN or OTPICS ($O(n \ log(n))$). Second, the stay-point detection step shrinks down the data point number to feed into the following clustering algorithm. To free the spatial constraint, PC-TC does not use stay-point detection. Therefore, PC-TC has more input data than other approaches. If the compared approaches should achieve a finer spatial granularity, they have to use a smaller distance parameter for the stay-point detection to produce a higher quantity of input data. Consequently, the compared approaches are expected to require more computation time. Our next evaluation confirms this inference.

As should be noted, the computation using approach 3 and 4 spend very long time when the cluster number is large (e.g., several thousands). Such large cluster numbers make no sense in practice, because human mobility is limited and one participant from the dataset hardly have visited thousands of places in two months. As the data sizes of the selected datasets after pre-processing are 3,648 and 3,984 data points on average respectively, we limit the cluster number to half of the input data size for DB and SC.

**Second evaluation.** Fig. 7 shows the results of our second evaluation. For the StudentLife dataset, the POI numbers by PC-TC decrease in a linear trend with the threshold values of $F_{vd}$ increasing from 0.1 to 0.9. Although the values of $\varepsilon$ (5, 25, 50 meters) can also moderate the POI number by DBSCAN, but the effect is small. As expected, the corresponding PLs change in the opposite direction. Interestingly enough, when PC-TC ($F_{vd}=0.9$) and DBSCAN ($\varepsilon=5$) produce the comparable level of POI number, the PL (0.82 in the median) by PC-TC is significantly higher than the one (0.77 in the median) by DBSCAN. We also find the same phenomena for the MDC dataset. This indicates that PC-TC has more potential for next place prediction applications. In terms of the computation time, DBSCAN requires more time when the value of $\varepsilon$ become smaller. This is caused by the increase of input data produced by stay-point detection with smaller values of $\varepsilon$.

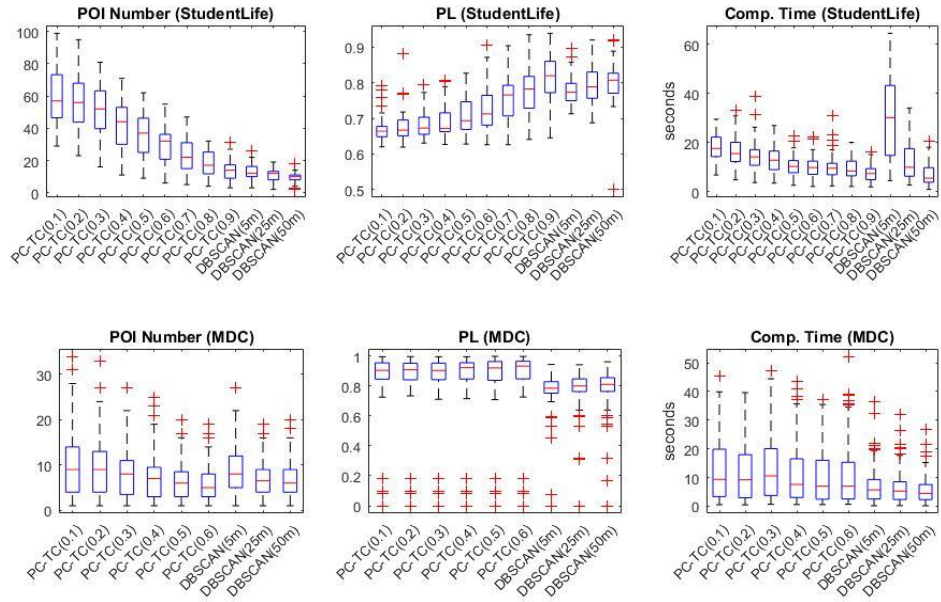

**Fig. 7.** The results of the second evaluation. The two left most graphs show the numbers of extracted POIs; the two middle graphs present the predictability limit of the POI sequences; and the two right most graphs represent the computation time.

## 5 Discussion

In comparison with other approaches, PC-TC uses temporal constraints instead of spatial constraints for POI extraction. The parameters required to define a POI are more intuitive and can be flexibly adjusted in practice. The results of our evaluations show that PC-TC perform better than compared approaches in terms of the number of extracted POIs and the predictability limit of the POI sequences. In other words, with the same level of POI numbers, the extracted POI sequences by PC-TC achieve higher predictability limits.

By following a new optimization objective, POI score, our approach has demonstrated its potential for next place prediction. Theoretically, POI score can be applied to any hierarchical clustering method. We use the complete-linkage algorithm in our approach as in [19]. However, we did not test other hierarchical clustering algorithms, which is one limitation of this work.

A second limitation applies to the datasets we used in our evaluation. The GPS location data in the StudentLife dataset were collected only every 10 or 20 minutes, which may miss mobility information and decreases the accuracy of duration estimation. The GPS location data in the MDC dataset only covers 2.4% of participants' time during the data collection period. We suggest that researchers working on data collection in this domain should follow the methods described in [30], which collects data points based on the change of a user's state instead of a time interval, to avoid losing much mobility information of users and to reduce the redundancy in the dataset.

Despite the objective evaluation, another limitation of this work is that the evaluation includes no subjective feedback, since we cannot reach the users who originally provided the data. For each user's data, only she or he can confirm the validity of detected places (e.g., a user often goes to a place near his office for smoking). For example, Montoliu et al. [10] employed a subjective method using the user-remembered places to evaluate the correctness and completeness of the extracted POIs.

## 6    Conclusion

In this paper, we proposed a novel POIs extraction approach called POIs clustering with temporal constraints (PC-TC) based on spatio-temporal data of human mobility. We introduced a new metric, POI-score, to evaluate clustering results and attain the optimal cluster number for hierarchical clustering. Taking advantage of the hierarchical information, PC-TC can generate POIs in two scales (i.e., global POIs and local POIs). We tested our approach using two publicly available datasets and compared it against four related approaches. In the StudentLife dataset, PC-TC extracts many more POIs, which shows the benefit of freeing the distance parameter. Compared to other approaches with the very similar number of extracted POIs, PC-TC obtains a higher predictability limit. Overall, the results of our experiment suggested that PC-TC is an efficient and practical approach for human mobility analysis and has an advantage in terms of next place prediction.

Potential future research includes: (1) investigating the impact of different parameter values on POI extraction results, (2) comparing different linkage algorithms (e.g., Ward's method [31]) in our approach, (3) exploring online hierarchical algorithms methods (e.g., BIRCH [32]) to allow long-term online analysis, as well as (4) conducting studies to provide subjective evaluation of our approach.